# Tropical Geometric Tools for Machine Learning: the **TML** package

David Barnhill    Ruriko Yoshida    Georgios Aliatimis
Keiji Miura

**Abstract**

In the last decade, developments in tropical geometry have provided a number of uses directly applicable to problems in statistical learning. The **TML** package is the first R package which contains a comprehensive set of tools and methods used for basic computations related to tropical convexity, visualization of tropically convex sets, as well as supervised and unsupervised learning models using the tropical metric under the max-plus algebra over the tropical projective torus. Primarily, the **TML** package employs a Hit and Run Markov chain Monte Carlo sampler in conjunction with the tropical metric as its main tool for statistical inference. In addition to basic computation and various applications of the tropical HAR sampler, we also focus on several supervised and unsupervised methods incorporated in the **TML** package including tropical principal component analysis, tropical logistic regression and tropical kernel density estimation.

## 1 Introduction

Tropical geometry is a relatively young field which involves examining the characteristics of geometric structures defined by the solution set of a series of a polynomial equations in max-plus, or tropical, algebra. Alternatively, tropical geometry can be described as the piecewise-linear analogue of classical geometry, as discussed in [10, 21]. In general, tropical geometry focuses on structures existing in the *tropical projective torus* defined as the quotient space $\mathbb{R}^e/\mathbb{R}\mathbf{1}$ which is isomorphic to $\mathbb{R}^{e-1}$. To date researchers have focused much attention on the theoretical underpinnings of tropical algebra and geometry (See [29, 10, 21, 14] for a thorough treatment).

As with any mathematical field, it is natural to examine how the ideas of tropical geometry can be used to solve various statistical problems. However, while methods are currently being developed for applied problems associated with tropical geometry, few comprehensive resources exist to handle such problems. In this article we attempt to fill this need by introducing the **TML** package [6], which provides a number of tropical statistical methods developed for use on problems associated with tropical geometry in the R programming language [26]. The **TML** package can be obtained through the Comprehensive R

Archive Network (CRAN) at https://cran.r-project.org/web/packages/TML/index.html.

## 1.1 Data science and tropical geometry

Statistical tools in data science are often classified in terms of supervised and unsupervised learning. In the case of supervised learning, data observations possess a dependent variable in the form of a label or quantity of interest which is used to train models in order to predict outcomes or classify unseen data based on those labels. Unsupervised learning is descriptive in nature where data observations possess no pre-determined response variable. The goal in unsupervised learning is to better understand the relationships between data observations or intuit the underlying structure of the data. A thorough yet concise summary with a number of open problems related to supervised and unsupervised learning as applied to tropical geometry can be found in [42]. We will leverage these terms in this article as well, for methods like tropical logistic regression (supervised learning), tropical principal component analysis (unsupervised learning), and tropical kernel density estimation (unsupervised learning). But we will also give significant focus to Markov chain Monte Carlo methods that do not necessarily fall exclusively in one class of machine learning but can be incorporated into supervised and unsupervised techniques or standalone for statistical inference computations.

While the research to develop tropical geometric statistical tools is nascent, there have been significant developments in a number of areas. One powerful tool in Euclidean space are Markov chain Monte Carlo (MCMC) samplers, which combine Monte Carlo sampling with Markov chains. Instead of Monte Carlo sampling according to a specific distribution, MCMC methods sample points by building a Markov chain that converges to a desired target distribution [17]. [44] introduced the first Markov chain Monte Carlo hit-and-run (HAR) sampling technique for use in the tropical projective torus. This method samples points uniformly from polytropes, which are tropical polytopes that are classically convex, as well as full-dimensional elements of any generic tropical simplices (see Definition 1.20). The authors also show how to sample points over the space of ultrametrics using line segments [44]. [4] extended this method to show how to sample points about a centroid with a given dispersion parameter in a method akin to Gaussian sampling in Euclidean space. These aforementioned HAR samplers feature prominently in several aspects of statistical learning as they have been applied in a number of settings. [43] employ HAR methods to execute non-parametric density estimation techniques over the space of ultrametrics which is known to be tropically convex [31]. [7] show how to employ HAR methods to estimate the volume of a tropical polytope. Being able to sample points this way paves the way for approaching statistical problems ranging from integral estimation to optimization in the setting of the tropical projective torus.

In supervised learning, [1] define the idea of tropical linear regression as the best approximation of a set of data observations using a tropical hyperplane. They then show the relationship of the tropical hyperplane approxima-tion with mean payoff games. [2] introduce a supervised classification method called *tropical logistic regression* for use over the space of rooted and equidistant phylogenetic trees. In this case, phylogenetic trees are defined in terms of ultrametrics and are classified into one of two *species trees* (see Section 1.4). Tropical analogues of other supervised methods exist such as tropical support vector machines [11]. Specifically, [11] define the notion of the tropical support vector machine (SVM) as a binary classification mechanism. Extensions of tropical SVMs as classifiers are exhibited in both [34] and [40].

Research into tropical unsupervised methods is also burgeoning. [41] introduce the tropical analogue of principal component analysis, (PCA) where the $n$-th order principal component can be represented as the best fit tropical polytope for a set of observations that are ultrametrics. It should be noted that the tropical PCA methods have been adapted to use HAR methods to find the best fit polytope. [5] introduce the tropical analogue of k-means and hierarchical clustering over the tropical projective torus as well as focusing on the space of ultrametrics.

## 1.2 The TML package

Availability of software tools in tropical geometry is plentiful when it comes to computer algebra. Singular is a computer algebra system for polynomial computations with emphasis on commutative algebra, algebraic geometry, and singularity theory which includes functionality for use in tropical geometry [9]. However, while Singular provides functionality for applications related to tropical geometry, there is no functionality for specific statistical methods. Similarly, polymake is a software used for research in polyhedral geometry which includes tropical geometry [13]. As with Singular, polymake focuses on the geometric and combinatorial analysis of polytopes. Additionally, polymake provides nice visualization options for tropical polytopes. The Open Source Computer Algebra Research (OSCAR) project provides an extensive corpus of tools by combining the functionality of several software environments including both Singular and polymakefor use in the Julia programming language [23].

While there exists a significant number of software programs that include a focus on tropical geometry as it relates to computer algebra, few resources exist specifically for statistical computation. And though the **algstat** package for the R programming language specifically focuses on algebra statistics, it provides no tools for the tropical case. In fact, there is no comprehensive suite of tropical statistical tools available. Nonetheless, some functionality exists in a piecemeal manner in the R programming language. Supervised and unsupervised learning methods are available in the **RTropical** package which makes use of tropical SVMs as well as tropical PCA over the space of phylogenetic trees [36]. Basic tropical arithmetic functions are available in the **tropical** package though this package has been archived.

In this article we introduce the **TML** which serves to provide a comprehensive suite of statistical tools applicable to tropical geometry for use in the R programming language. The package consists of functions and methods rang-ing from basic tropical arithmetic and linear algebra functions to more complex supervised and unsupervised tropical machine learning techniques. The **TML** package is distributed on the Comprehensive R Archive Network (CRAN) with version control managed through Git on Github (https://github.com/barnhilldave/TML).

The organization of this article is as follows. In Section 1.3 we offer the essential elements of tropical geometry that provide the background for the methods introduced in the **TML** package. In Section 2 we illustrate basic loading and operations of the **TML** package as well as visualization methods. Because MCMC methods feature prominently in the **TML** package, Section 3 focuses on the tropical HAR sampler developed in [44] called *tropical HAR with extrapolation* and illustrates its usage. Section 4 follows with examples for the prominent methods of statistical inference in the **TML** package. These applications include volume estimation of tropical polytopes, tropical logistic regression, tropical PCA, and tropical kernel density estimation. We finish the article with concluding remarks and potential future developments.

## 1.3 Tropical Basics

In this section we provide the necessary tropical geometric background, notation, and terminology as it relates to functions in the **TML** package. This section is divided into two subsections. The first section focuses on tropical analogues of arithmetic and linear algebraic operations. The second subsection defines several essential concepts associated with tropical polyhedral geometry.

### 1.3.1 Tropical arithmetic

In general, research in tropical geometry studies the properties space defined by the *tropical semiring* represented by the triplet $(\mathbb{R} \cup \{-\infty\}, \oplus, \odot)$. Here, classical addition is replaced by *max* ($\oplus$) and classical multiplication is replaced by classical addition ($\odot$) [10, 21]. This space is known as the *tropical projective torus* represented by $\mathbb{R}^e/\mathbb{R}\mathbf{1}$, where $\mathbf{1} := (1, 1, \ldots, 1)$ is the vector with all ones in $\mathbb{R}^e$. This requires that if $v := (v_1, \ldots, v_e) \in \mathbb{R}^e/\mathbb{R}\mathbf{1}$,

$$(v_1 + c, \ldots, v_e + c) = (v_1, \ldots, v_e) = v. \tag{1}$$

For an extensive treatment, see [14] and [21].

**Definition 1.1** (Tropical Arithmetic Operations). *Under the tropical semiring with the tropical, or* max-plus *algebra* $(\mathbb{R} \cup \{-\infty\}, \oplus, \odot)$*, we have the arithmetic operations of addition and multiplication defined as:*

$$x \oplus y := \max\{x, y\}, \quad x \odot y := x + y \quad \text{where } x, y \in \mathbb{R} \cup \{-\infty\}.$$

*Here $-\infty$ is the identity element under addition $\oplus$ and $0$ is the identity element under multiplication $\odot$.*

*We may also define the tropical semiring with the* min-plus algebra *(*$\mathbb{R} \cup \{\infty\}, \boxplus, \odot$*), where arithmetic operations of addition and multiplication defined as:*

$$x \boxplus y := \min\{x, y\}, \quad x \odot y := x + y \quad \text{where } x, y \in \mathbb{R} \cup \{\infty\}.$$

**Definition 1.2** (Tropical Scalar Multiplication and Vector Addition)**.** *For any $x, y$ $\in \mathbb{R} \cup \{-\infty\}$ and for any $v = (v_1, \ldots, v_e)$, $w = (w_1, \ldots, w_e) \in (\mathbb{R} \cup \{-\infty\})^e$, we have tropical scalar multiplication and tropical vector addition defined as:*

$$x \odot v \oplus y \odot w := (\max\{x + v_1, y + w_1\}, \ldots, \max\{x + v_e, y + w_e\}).$$

**Definition 1.3** (Generalized Hilbert Projective Metric)**.** *For any tropical points $v := (v_1, \ldots, v_e)$, $w := (w_1, \ldots, w_e) \in \mathbb{R}^e/\mathbb{R}\mathbf{1}$ where $[e] := \{1, \ldots, e\}$, the* trop*ical distance (also known as* tropical metric*) $d_{\text{tr}}$ between $v$ and $w$ is defined as:*

$$d_{\text{tr}}(v, w) := \max_{i \in [e]}\{v_i - w_i\} - \min_{i \in [e]}\{v_i - w_i\}.$$

For any two points $v, w \in \mathbb{R}^e/\mathbb{R}\mathbf{1}$, the tropical distance between $v$ and $w$ assumes Equation 1 holds. Otherwise the tropical distance is not a metric.

**Definition 1.4** (Tropical Determinant [14])**.** *Let $w$ be a positive integer. For any square tropical matrix $B$ of size $w \times w$ with entries in $\mathbb{R} \cup \{-\infty\}$, we say the tropical determinant of $B$ as follows:*

$$tdet(B) := \max_{\sigma \in S_w}\{B_{\sigma(1),1} + B_{\sigma(2),2} + \ldots + B_{\sigma(w),w}\}, \quad (2)$$

*where we denote the $(i,j)$-th entry of $B$ as $B_{i,j}$ and $S_w$ represents every permutation of $[w] := \{1, \ldots, w\}$.*

### 1.3.2 Tropical geometric structures

The definitions that follow describe and clarify characteristics of polytopes and hypterplanes in the tropical projective torus.

**Definition 1.5** (Tropical line segment from [21])**.** *Given two points $u, v$, a tropical line segment between $u, v$ denoted as $\Gamma_{u,v}$, consists of the concatenation of at most $e - 1$ Euclidean line segments. A point in the collection of points* **b***, defining the end points of each Euclidean line segment is called a* bend point *of $\Gamma_{u,v}$. Including $u$ and $v$, $\Gamma_{u,v}$ consists of at most $e$ bend points. We show how to compute the set* **b** *in (3).*

[21] show how to construct a tropical line segment between two vectors in the proof of their Proposition 5.2.5. For a pair of vectors $u =: (u_1, \ldots, u_e), v := (v_1, \ldots, v_e) \in \mathbb{R}^e/\mathbb{R}\mathbf{1}$, the tropical line segment can be constructed as follows: Without loss of generality, assume that $(v_1 - u_1) \geq \ldots \geq (v_{e-1} - u_{e-1}) \geq$

$(v_e - u_e) = 0$ over the tropical projective torus $\mathbb{R}^e/\mathbb{R}\mathbf{1}$ once the coordinates of $v - u$ have been permuted. The tropical line segment $\Gamma_{u,v}$ from $v$ to $u$ is

$$\begin{cases} (v_e - u_e) \odot u \oplus v &= v \\ (v_{e-1} - u_{e-1}) \odot u \oplus v &= (v_1, v_2, v_3, \ldots, v_{e-1}, v_{e-1} - u_{e-1} + u_e) \\ &\vdots \\ (v_2 - u_2) \odot u \oplus v &= (v_1, v_2, v_2 - u_2 + u_3, \ldots, v_2 - u_2 + u_e) \\ (v_1 - u_1) \odot u \oplus v &= u. \end{cases} \quad (3)$$

**Definition 1.6** (Tropical Polytopes [14]). *Suppose we have $S \subset \mathbb{R}^e/\mathbb{R}\mathbf{1}$. If*

$$x \odot v \oplus y \odot w \in S$$

*for any $x, y \in \mathbb{R}$ and for any $v, w \in S$, then $S$ is called* tropically convex. *Suppose $V = \{v^1, \ldots, v^s\} \subset \mathbb{R}^e/\mathbb{R}\mathbf{1}$. The smallest tropically convex subset containing $V$ is called the* tropical convex hull *or* tropical polytope *of $V$ which can be written as the set of all tropical linear combinations of $V$*

$$\text{tconv}(V) = \{a_1 \odot v^1 \oplus a_2 \odot v^2 \oplus \cdots \oplus a_s \odot v^s \mid a_1, \ldots, a_s \in \mathbb{R}\}.$$

*The smallest subset $V' \subseteq V$ such that*

$$\text{tconv}(V') = \text{tconv}(V)$$

*is called a minimum, or generating set, with $|V'|$ being the cardinality of $V'$. For $P = tconv(V')$ the boundary of $P$ is denoted $\partial P$.*

**Remark 1.7.** *A tropical polytope of two points $u, v \in \mathbb{R}^e/\mathbb{R}\mathbf{1}$ is called a tropical line segment, denoted $\Gamma_{u,v}$.*

We use the term *max-tropical polytope* if a tropical polytope is defined in terms of the max-plus algebra. Conversely, we use the term *min-tropical polytope* if a tropical polytope is defined in terms of the min-plus algebra. When the context is clear we will just use the term tropical polytope.

**Definition 1.8** (Polytropes [15]). *A classically convex tropical polytope is called a* polytrope.

**Definition 1.9** (Tropical Simplex). *A tropical simplex is a tropical polytope that possess a minimum vertex, or generating, set $V'$ such that $|V'| = e$. A tropical simplex is denoted $P_\Delta$. All polytropes are tropical simplices. The converse is not true.*

An important type of polytrope is a *tropical ball*. A tropical ball is the analogue of a Euclidean ball which is defined as

$$B_r(x) = \{y \in \mathbb{R}^e \mid ||x - y||_2 \leq r\}$$

and indicates the set of all points falling within a distance $r$ of a point $x$ where distance is calculated by the $L_2$-norm. In the case of a tropical ball, distance is defined in terms of the tropical metric shown in Definition 1.3.

**Definition 1.10** (Tropical Ball). *A tropical ball, $B_l(x_0)$, around $x_0 \in \mathbb{R}^e/\mathbb{R}\mathbf{1}$ with a radius $l > 0$ is defined as follows:*

$$B_l(x_0) = \{y \in \mathbb{R}^e/\mathbb{R}\mathbf{1} \,|\, d_{\text{tr}}(x_0, y) \leq l\}.$$

*The minimum generating set $V'$ of a tropical ball consists of exactly $e$ vertices in which case a tropical ball is a tropical simplex. Figure 1 provides the generic structure of a tropical ball.*

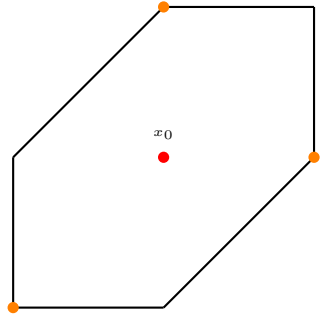

Figure 1: Tropical ball, $B_l(x_0) \in \mathbb{R}^3/\mathbb{R}\mathbf{1}$ with radius $l$. Center point is indicated in red with the generating set $V'$ shown in orange (Credit: [4]).

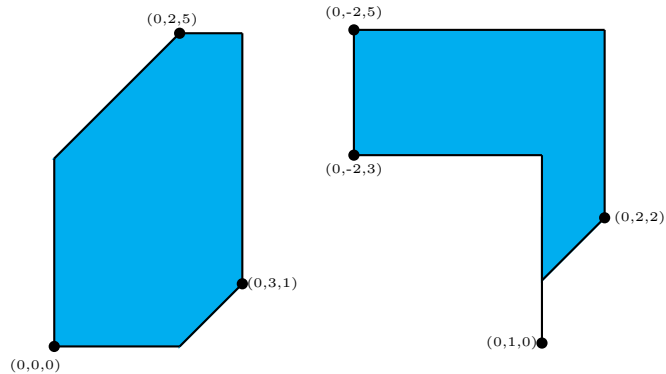

Figure 2: Tropical polytopes in $\mathbb{R}^3/\mathbb{R}\mathbf{1}$. The tropical polytope on the left is a polytrope and therefore a tropical simplex (Credit: [4]).

In many situations, we are interested in the projection of a point onto a tropical polytope. This projection can be represented by using Formula 5.2.3 in [21] and is shown below.

**Definition 1.11** (Tropical Projection). *Let $V := \{v^1, \ldots, v^s\} \subset \mathbb{R}^e/\mathbb{R}\mathbf{1}$ and let $\mathcal{P} = tconv(v^1, \ldots, v^s) \subseteq \mathbb{R}^e/\mathbb{R}\mathbf{1}$ be a tropical polytope with its vertex set $V$. For*

$\mathbf{x} \in \mathbb{R}^e/\mathbb{R}\mathbf{1}$, let

$$\pi_{\mathcal{P}}(\mathbf{x}) := \bigoplus_{l=1}^{s} \lambda_l \odot v^l, \quad \text{where} \quad \lambda_l = \min\{\mathbf{x} - v^l\}. \tag{4}$$

Then

$$d_{\text{tr}}(\mathbf{x}, \pi_{\mathcal{P}}(\mathbf{x})) \leq d_{\text{tr}}(\mathbf{x}, \mathbf{y})$$

for all $\mathbf{y} \in \mathcal{P}$.

Now we turn our attention to definitions which will exhibit the relationship between min-tropical hyperplanes and max-tropical polytopes.

**Definition 1.12** (Tropical Hyperplane [29]). *For any $\omega := (\omega_1, \ldots, \omega_e) \in \mathbb{R}^e/\mathbb{R}\mathbf{1}$, the max-tropical hyperplane defined by $\omega$, denoted as $H_\omega^{\max}$, is the set of points $x \in \mathbb{R}^e/\mathbb{R}\mathbf{1}$ such that*

$$\max_{i \in [e]} \left\{ \omega_i + x_i \right\} \tag{5}$$

*is attained at least twice. Similarly, a min-tropical hyperplane denoted as $H_\omega^{\min}$, is the set of points $x \in \mathbb{R}^e/\mathbb{R}\mathbf{1}$ such that*

$$\min_{i \in [e]} \left\{ \omega_i + x_i \right\} \tag{6}$$

*is attained at least twice. If it is clear from context, we simply denote $H_\omega$ as a tropical hyperplane in terms of the min-plus or max-plus algebra where $\omega$ is the* normal vector *of $H_\omega$. The point $-\omega$ represents a point contained in $H_\omega$ where the maximum or minimum is attained e times. This point is called the* apex *of $H_\omega$.*

**Definition 1.13** (Sectors from Section 5.2 in [14]). *Every tropical hyperplane, $H_\omega$, divides the tropical projective torus, $\mathbb{R}^e/\mathbb{R}\mathbf{1}$ into e connected components, which are* open sectors

$$S_\omega^i := \{x \in \mathbb{R}^e/\mathbb{R}\mathbf{1} \,|\, \omega_i + x_i > \omega_j + x_j, \forall j \neq i\}, \ i = [e].$$

*These* closed sectors *are defined as*

$$\overline{S}_\omega^i := \{x \in \mathbb{R}^e/\mathbb{R}\mathbf{1} \,|\, \omega_i + x_i \geq \omega_j + x_j, \forall j \neq i\}, \ i = [e].$$

**Lemma 1.14** (Distance to a Tropical Hyperplane $H_\omega$ [11]). *The tropical distance from a point $v \in \mathbb{R}^e/\mathbb{R}\mathbf{1}$ to a max-tropical hyperplane $H_0^{max}$ where $\omega$ represents the normal vector of zeros, is given by the difference between the maximum and second maximum of $v$. That is*

$$d_{tr}(H_0, v) = \max(v) - 2nd \max(v). \tag{7}$$

*For a min-tropical hyperplane $H_0^{min}$, the tropical distance from a point $v$ is*

$$d_{tr}(H_0, v) = 2nd \min(v) - \min(v). \tag{8}$$

*For a generic tropical hyperplane $H_\omega$,*

$$d_{tr}(H_\omega, v) = d_{tr}(H_0, v + \omega). \tag{9}$$

Figure 3 illustrates the construction of $H_\omega^{max}$ with apex at $v = (0, 4, 7)$ and $\omega = -v$. Additionally, we observe the point $u = (0, 3, 1)$ in relation to $H_\omega^{max}$ along with each sector $\bar{S}_\omega^i$ as shown in Definition 1.13.

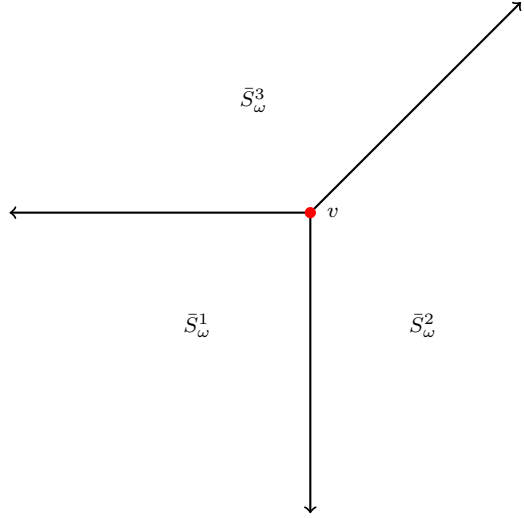

Figure 3: Generic tropical hyperplane $H_\omega^{max}$ in $\mathbb{R}^3/\mathbb{R}\mathbf{1}$ with associated closed sectors. The red point represents the apex, $v$ with $\omega = -v$ (Credit: [4]).

**Definition 1.15** (Tropical Hyperplane Arrangements). *For a given set of points, $V = \{v^1, \ldots, v^s\}$, tropical hyperplanes with apices at each $v^i \in V$ represent the* tropical hyperplane arrangement *of $V$, $\mathcal{A}(V)$, where*

$$\mathcal{A}(V) := \{H_{-v^1}, \ldots, H_{-v^s}\}.$$

*If we consider a collection of tropical hyperplanes defined in terms of the max-plus algebra, we call this arrangement a* max-tropical hyperplane arrangement *denoted $\mathcal{A}^{\max}(V)$. Likewise, considering tropical hyperplanes defined in terms of the min-plus algebra is called a* min-tropical hyperplane arrangement *denoted $\mathcal{A}^{\min}(V)$.*

**Definition 1.16** (Cells). *For a given hyperplane arrangement, $\mathcal{A}(V)$, a* cell *is defined as the intersection of a finite number of closed sectors. Cells may be* bounded *or* unbounded. *Bounded cells are polytropes.*

**Definition 1.17** (Bounded Subcomplex [10]). *For a vertex set, $V$, $\mathcal{A}(V)$ defines a collection of bounded and unbounded cells which is known as a* cell decomposition. *The union of bounded cells defines the* bounded subcomplex, $\mathcal{K}(V)$.

**Theorem 1.18** (Corollary 6.17 from [14]). *A* max-*tropical polytope, $P$, is the union of cells in $\mathcal{K}(V)$ of the* cell decomposition *of the tropical projective torus induced by $\mathcal{A}^{\min}(V)$.*

Theorem 1.18 describes $\mathcal{K}(V)$ as a collection of bounded cells induced by some $\mathcal{A}(V)$. Figure 4 represents a tropical polytope in terms of its vertex set (left) and hyperplane arrangement (right).

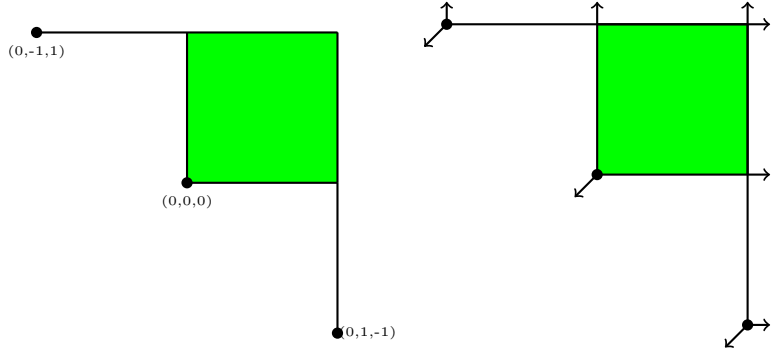

Figure 4: Tropical polytope expressed in terms of its vertex set $V$ (left) and $\mathcal{A}^{\min}(V)$ (right). In the right figure, pseudovertices not belonging to $V$ are defined by the intersection of min-tropical hyperplanes with apices at each $v^i \in V$ (Credit: [4]).

Throughout this paper we are interested in sampling the union of $(e-1)$-dimensional polytropes belonging to $\mathcal{K}(V)$. The union of $(e-1)$-dimensional polytropes is described in the following definition.

**Definition 1.19** (Dimension of a Tropical Polytope). *The dimension of a tropical polytope, $P \in \mathbb{R}^e/\mathbb{R}\mathbf{1}$, is defined by the bounded cell of maximal dimension in $\mathcal{C}_P$ and is denoted as $\dim(P)$.*

**Definition 1.20** (*i-trunk* and *i-tentacles* (Definition 2.1 in [20])). *Let $P$ be a tropical polytope and let $i \in [e-1]$ where $[e-1] = \{1, \ldots, e-1\}$. Let $\mathcal{F}_P$ be the family of relatively open tropical polytopes in $\mathcal{C}_P$. For any $T \in \mathcal{F}_P$, $T$ is called an* i-tentacle element *of $\mathcal{F}_P$ if it is not contained in the closure of any $(i+1)$-dimensional tropical polytope in $\mathcal{F}_P$ where the dimension of $T$ less than or equal to $i$. The* i-trunk *of $P$, is defined as*

$$Tr_i(P) := \bigcup \left\{ F \in \mathcal{F}_P : \exists\, G \in \mathcal{F}_P \text{ with } \dim(G) \geq i \text{ such that } F \subseteq G \right\}$$

*where $\dim(G)$ is the dimension of $G \subset \mathcal{F}_P$. The $Tr_i(P)$ represents the portion of the $\mathcal{K}(V)$ with $(i-1)$-tentacles removed. The minimum enclosing ball containing containing only $Tr_i(P) \subseteq P$ is denoted $B_k(Tr_i(P))$.*

**Example 1.21.** *Consider the tropical polytope, $P = \{(0,0,0), (0,-1,1), (0,2,2), (0,1,-1)\}$. The $Tr_2(P)$ is the gray portion shown in Figure 5.*

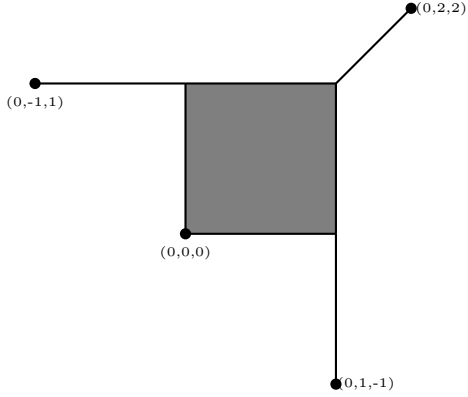

Figure 5: A tropical polytope, $P$, in $\mathbb{R}^3/\mathbb{R}\mathbf{1}$ defined by four vertices. The $Tr_2(P)$ is the portion in gray (Credit: [7]).

For many statistical problems, a first step in statistical inference is finding a centroid for some given data. This is no less true when handling data in the tropical projective torus. To that end, we concern ourselves with finding the Fermat-Weber point in the tropical projective torus, or simply the tropical Fermat-Weber point.

**Definition 1.22** (Tropical Fermat-Weber Points (see [19]))**.** *For a given set of points $V = \{v^1, \ldots, v^s\}$ in a metric space $\mathcal{M}$, the Fermat-Weber point for the set $V$ is*

$$\arg\min_y \sum_{i=1}^s d(y, v_i), \tag{10}$$

*where $d(.)$ represents the function defining a distance metric in $\mathcal{M}$ and the point $y \in \mathcal{M}$ represents a centroid. A* tropical Fermat-Weber point *is similarly defined but replacing $d(.)$ with the tropical metric $d_{tr}(.)$*

$$\arg\min_y \sum_{i=1}^s d_{tr}(y, v_i). \tag{11}$$

**Remark 1.23.** *The tropical Fermat-Weber point is not guaranteed to be unique (see [19]).*

In this paper we will introduce a number of ways to obtain a tropical FW point and then apply the resulting point, or centroid, to problems of statistical inference.

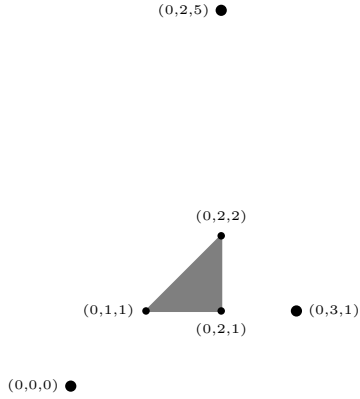

Figure 6: Fermat–Weber region defined by three points. Points in the gray triangle satisfies (11) [19, 5].

## 1.4 Tropical Geometry and Phylogenetic Trees

One area of interest where tropical geometry is directly applicable is in the biological science of phylogenomics. Phylogenomics is a discipline focusing on reconstructing the evolutionary history of organisms. One method of representing this evolutionary history is through the use of phylogenetic trees. Phylogenetic trees are data structures representing the evolution of genes from related, and usually present-day, species or some other taxa. In terms of graph theory, phylogenetic trees represent rooted out-trees consisting of a root node, unlabeled internal nodes, and external leaf nodes. The root node of the tree represents a common evolutionary ancestor among several taxa, internal nodes represent speciation events over time, and the external, leaf, nodes represent the present-day taxa.

Of particular interest, as it relates to tropical geometry, are those phylogenetic trees that are *equidistant* trees. Equidistant trees are trees where the distance from the root node to each leaf node is the same. It is shown in [8] that equidistant trees can be represented as ultrametrics which are described in Definition 1.24 and illustrated in Example 1.25.

**Definition 1.24** (Ultrametric)**.** *Let $[m] := \{1, \ldots, m\}$ and define the distance function $d : [m] \times [m] \to \mathbb{R}$ to be a metric over $[m]$. Then if*

$$\max\{d(i,j), d(i,k), d(j,k)\}$$

*is attained at least twice for any $i, j, k \in [m]$, $d$ is an* ultrametric*.*

**Example 1.25.** *Suppose $m = 3$. Let $d$ be a metric on $[m] := \{1, 2, 3\}$ such that*

$$d(1,2) = 2,\, d(1,3) = 2,\, d(2,3) = 1.$$

*Since the maximum is achieved twice, $d$ is an ultrametric.*

Specifically, for any equidistant tree $T \in \mathcal{U}_m$ where $\mathcal{U}_m$ represents the space of ultrametrics on $m$ leaves, the vector representing the pairwise distances between leaf nodes is an ultrametric. Further, it was shown in [32] that $\mathcal{U}_m$ is a tropical Grassmanian and is therefore tropically convex. This affords us the opportunity to apply statistical methods based on tropical geometry to the space of equidistant phylogenetic trees. We will leverage this characteristic of phylogenetic trees throughout this paper by applying the methods of the **TML** package to equidistant phylogenetic trees.

### 1.4.1 The Coalescent Model

One way to evaluate how well supervised and unsupervised learning techniques perform when empirical data is unavailable is by using simulated data sets. One method to obtain simulated data is by using the *coalescent model*. The coalescent model was first described in [16] and, for a given sample of taxa, represents a stochastic genealogical process illustrating coalescing events which shows the ancestral lineage of those related taxa. The coalescent model uses species tree to represent the overall lineage of related species with phylogenetic trees showing the lineage of specific genes of the related species in the species tree. We can think of phylogenetic trees emanating from a species tree [27]. For a thorough treatment of the coalescent model see [35]. Using the coalescent model we can simulate samples of phylogenetic trees that come from specific species trees.

The coalescent is a common model used in a wide range of software. For the simulated data used in this article and included in the **TML** package, we employ the software called Mesquite [22]. Mesquite takes the arguments of species depth, $SD$, and effective population, denoted $N_e$. The $SD$ indicates the number of epochs between the common ancestor (root node) and taxa of the present day (leaf nodes). We define the gene trees in terms of the ratio

$$R = \frac{SD}{N_e}.$$

If we set $N_e = 100000$ and $SD = 500000$, then we have $R = 5$. Notably, the Mesquite software does not represent equidistant trees as ultrametrics. However, the output of equidistant trees taken from Mesquite can be manipulated into ultrametric form using tools from the **phytools** R package [23]. A total of twelve data sets, each representing 1000 simulated phylogenetic trees on ten leaves with $R$ taking the values of 0.25, 0.5, 1, 2, 5, and 10, coming from two different species trees were constructed from Mesquite are included in the **TML** package. Using **phytools**, these twelve data sets were manipulated into ultrametrics and are included as `Sim_Trees1` for the six datasets coming from species tree one and `Sim_Trees2` from those coming from species tree two.

# 2 Basic operations

This section describes the basic functionality of **TML** and how to execute some simple computations.

## 2.1 Loading TML and basic operations

The **TML** package is loaded as all R packages are loaded:

```
R> library(``TML'')
```

Once **TML** is loaded all functionality is available. We note that functions are based on linear algebraic operations in terms of this max-plus algebra. Most inputs into functions in **TML** involve vectors or matrices. For example if we wish to find the tropical distance between two points in $\mathbb{R}^3/\mathbb{R}\mathbf{1}$ we input the following:

```
R> u <- (0,1,2)
R> v <- (0,4,7)
R> trop.dist(u,v)

[1] 5
```

As is shown in Equation 1, adding a scalar $c$ to each element of the vector in $\mathbb{R}^e/\mathbb{R}\mathbf{1}$ represents the same vector. This means that for a given vector, we can normalize the vector by adding the additive inverse of the value of any element of the vector. For example:

```
R> u <- (2,3,4)
R> normaliz.vector(u)

[1] 0 1 2
```

In practice we can choose any element of the vector but by convention **TML** forces the scalar value in the first position to be zero. As we observe, if we instead have unnormalized vectors we still obtain the same tropical distance between points obviating the need to normalize vectors prior to using the function.

```
R> u <- (2,3,4)
R> v <- (3,7,10)
R> trop.dist(u,v)

[1] 5
```

We can also apply a normalizing operation to a set of vectors. In general, we consider a set of vectors as defining a tropical polytope where each row vector consists of a point in the polytope. Importantly, the collection of points need not be the minimum generating set of the tropical polytope. The code snippet below illustrates the use of the `normaliz.polytope()` function on a set of vectors that defines a tropical polytope. We can say that both matrices represent the same polytope by using Equation (1).

```
R> P<-matrix(c(3,3,3,4,6,9,2,5,3),3,3,TRUE)
R> P

     [,1] [,2] [,3]
[1,]    3    3    3
[2,]    4    6    9
[3,]    2    5    3

R> normaliz.polytope(P)

     [,1] [,2] [,3]
[1,]    0    0    0
[2,]    0    2    5
[3,]    0    3    1
```

As with classical linear algebra we can find the tropical analogue of the determinant of a matrix as shown in Equation 1.4. This is equivalent to a linear assignment problem optimization problem as discussed in [14]. Again, the input is a matrix where rows are the points in $\mathbb{R}/\mathbb{R}\mathbf{1}$.

```
R> P <- matrix(c(0,0,0,0,2,5,0,3,1),3,3,TRUE)
R> tdets(P)

[[1]]
[1] 8

[[2]]
     [,1] [,2] [,3]
[1,]    0    0    0
[2,]    0    3    1
[3,]    0    2    5
```

This output comes in the form of a list where the first element represents the value of the tropical determinant and the second element is a matrix of the same points reordered such that the elements of each row vector contributing to the value of the determinant are on the diagonal.

It is common to work with phylogenetic trees, but the statistical methods of tropical geometry use vectorized input. This vector consists of components that express the pairwise distances between the $m$ leaves as the sum of branch lengths connecting those leaves. Consequently, the vector has dimension $e = \binom{m}{2}$. The following example illustrates how a tree with 4 leaves is converted to a vector in $\mathbb{R}^6$.

```
R> tree <- ape::read.tree(text='((A:1, B:1):2, (C:1, D:1):2);')
R> tree.to.vector(tree,normalization = F)

[1] 2 6 6 6 6 2
```

## 2.2 Tropical line segments, hyperplanes, polytopes, and projections

This section focuses on the functions in **TML** related to tropical geometric structures such as tropical line segments, polytopes, hyperplanes, and projections with associated computations.

Using Equation 3 we can construct the tropical line segment by calculating the associated bend points.

```
R> u <- (0,1,2)
R> v <- (0,4,7)
R> TLineSeg(u,v)

[[1]]
[1] 0 4 7

[[2]]
[1] 3 4 7

[[3]]
[1] 5 6 7
```

The previous example represents a line segment from the vector $v = (0, 4, 7)$ to $u = (0, 1, 2)$. The output is a list of vectors where each vector represents a bend point or end point of the tropical line segment. Note that in this case the second and third bend points are not normalized (i.e., the first value of each vector equal to zero). This can easily be accomplished using the `lapply()` function in conjunction with the `normaliz.vector()` function from **TML**.

```
R> u <- (0,1,2)
R> v <- (0,4,7)
R> lapply(TLineSeg(u,v),function(x) normaliz.vector(x))

[[1]]
[1] 0 4 7

[[2]]
[1] 0 1 4

[[3]]
[1] 0 1 2
```

It should be noted also that for two vectors in $\mathbb{R}^e/\mathbb{R}\mathbf{1}$ the `TLineSeg()` function output list will consist of $e$ bend points but some bend points may be redundant. This is an indication that the line segment consists of fewer than $e$ bend points.

Tropical hyperplanes, as defined in Equation 5 are defined simply by the point in the tropical projective torus that serves as the hyperplane apex. The **TML** package provides functions to visualize tropical hyperplanes in two and three dimensions as well as functions to measure the tropical distance from a point in the tropical projective torus to the nearest point on the hyperplane. Three dimensional tropical hyperplanes are also able to be visualized.

```
R> D<-t(as.matrix(c(0,0,0)))
R> E<-t(as.matrix(c(0,0,0,0)))
R> di<-4
R> mi<- -5
R> ma<- 5
R> hyper3d_max(D, di, mi, ma, plt=TRUE)
R> hyper3d_min(D, di, mi, ma, plt=TRUE)
R> hyper3d_max(E, di, mi, ma, plt=TRUE)
R> hyper3d_min(E, di, mi, ma, plt=TRUE)
```

Visualizations from the output of the previous code is shown in Figure 7. Max-tropical hyperplanes in both two dimensions and three dimensions are shown on the left with min-tropical hyperplanes shown on the right.

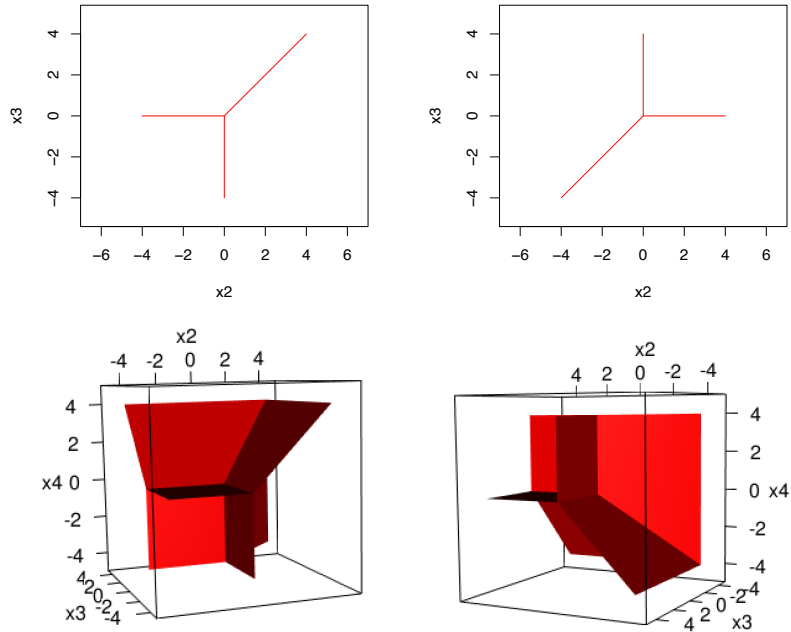

Figure 7: Max-tropical (left) and min-tropical hyperplanes (right) in two and three dimensions using the `hyper3d_max()` and `hyper3d_min()` functions.

The statistical methods introduced in this article utilize the tropical distance from a point to a tropical hyperplane. The distance from a point to a tropical hyperplane is shown in Lemma 1.14 for both the max-tropical and min-tropical case. The functions `trop.dist.hyp_max()` and `trop.dist.hyp_min()` take two inputs: the normal vector associated with the apex of the tropical hyperplane and any generic point in the tropical projective torus.

```
R> O <- c(0,-1,-1)
R> x0 <- c(0,-2,-8)
R> trop.dist.hyp_max(O,x0)

[1] 3
```

```
R> O <- c(0,-1,-1)
R> x0 <- c(0,-2,-8)
R> trop.dist.hyp_min(O,x0)

[1] 6
```

Recall from Definition 1.12 the normal vector is the same as changing the sign of the point in the tropical projective torus representing the apex of the tropical hyperplane. Figure 8 shows a point on the tropical hyperplane that is closest to the point $(0, -2, -8)$. The tropical distance to this point is shown in the code above. Notably, unlike Euclidean using a Euclidean distance, the point on the tropical hyperplane that is closest to the point of interest may not be unique.

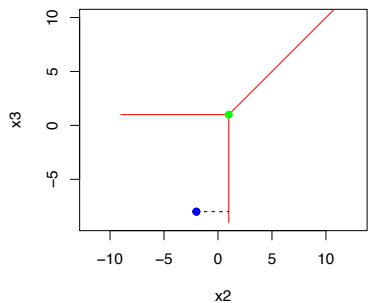

Figure 8: Tropical distance from a point $(0, -2, -8)$ (blue) to the max-tropical hyperplane defined by the normal vector $\omega = (0, -1, -1)$. Note that $\omega$ corresponds with the apex of the tropical hyperplane $(0, 1, 1)$ (green).

We now turn our attention to tropical polytopes as tropical polytopes are the primary geometric structures that are used in most functions in the **TML** package. The primary focus here is to illustrate how we visualize different tropical polytopes. Visualization is combinatorially challenging even in lower dimensions. Here we show a couple of examples of visualizations of two-dimensional and three-dimensional tropical polytopes. As stated in Section 1.3, a tropical ball is an important polytrope in tropical geometry. The function `Trop_ball()` allows us to render a two- or three-dimensional tropical ball.

```
R> v <- c(0,0,0)
R> d <- 2
R> Trop_ball(c(0,0,0),d,a=1,cls='white',fil=TRUE,cent.col='red')
R> Trop_ball(c(0,0,0,0),d,a=.5,cls='lightblue',
      fil=TRUE,cent.col='red')
```

The code above takes several inputs the first being the center of the tropical ball, the radius of the tropical ball in terms of tropical distance, and several other inputs involving the transparency and color options. Figure 9 provides the output of two tropical balls from the example above in two dimensions (left) and three dimensions (right).

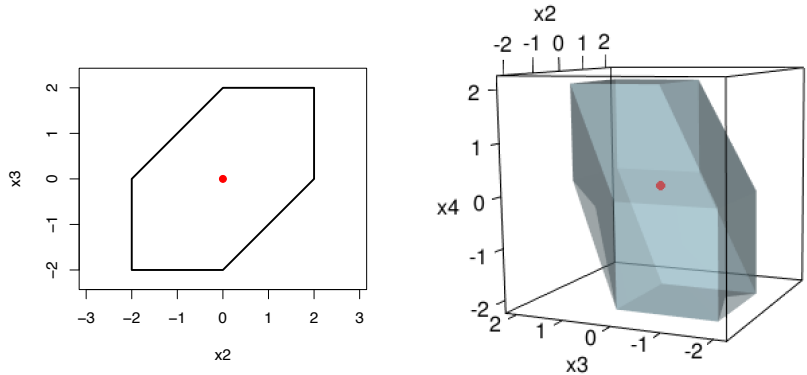

Figure 9: Tropical balls in two and three dimensions each with a radius of two, using the `Trop_ball()` function. The red point in each figure represents the center point. Black points represent the vertices of the tropical polytope.

Visualization of generic tropical polytopes can be accomplished in two or three dimensions using the `draw.tpolytope.2d()` and `draw.tpolytope.3d()` functions. These functions take three inputs: a matrix of points in the tropical projective torus, a color argument for the polytope itself and a color argument for the vertices of the polytope.

```
R> P <- matrix(c(0,-2,2,0,-2,5,0,2,1,0,1,-1),4,3,TRUE)
R> c <- 'darkgreen'
R> cc <- 'black'
R> draw.tpolytope.2d(P,c,cc)
```

```
R> P <- matrix(c(0,0,0,0,0,1,2,5,0,1,3,1,0,2,5,10),4,4,TRUE)
R> c <- 'darkgreen'
R> cc <- 'black'
R> draw.tpolytope.3d(P,c,cc)
```

Figures 10 and 11 show the output from the previous code. Note that in each case the function draws line segments between each vertex (and more points in the three-dimensional case) in the polytope. This is due to the combinatorial challenge with forming this polytopes. The tropical polytopes themselves are defined by the boundary.

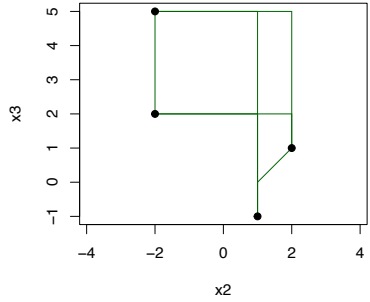

Figure 10: 2-D rendering of a tropical polytope using the `draw.tpolytope.2d()` function.

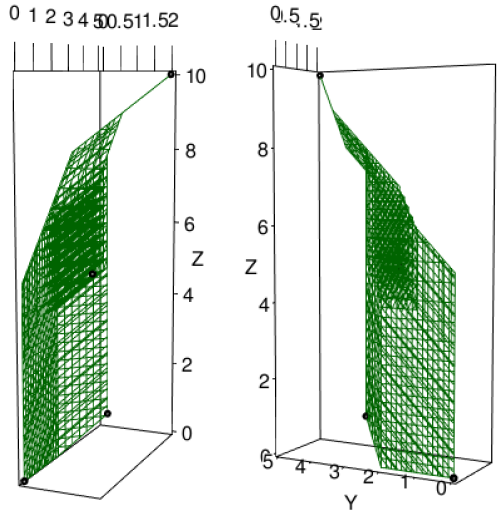

Figure 11: 3-D rendering of a tropical polytope using the `draw.tpolytope.3d()` function.

The last specific function we address in this section is the `project_pi()` function. The projection of a point $x$ onto a tropical polytope, $\mathcal{P}$, denoted $\pi(x)$ is the point in $\mathcal{P}$ that is closest to $x$ in terms of tropical distance. It must be noted that unlike Euclidean distance, tropical projections are not necessarily unique. In fact oftentimes there is an interval of points satisfying satisfying Equation 4. For a thorough discussion of projections see [4].

```
R> P <- matrix(c(0,0,0,0,2,5,0,3,1),3,3,TRUE)
R> x <- c(0,6,2)
R> project_pi(P,x)

[1] 0 3 2
```

The output above provides one of perhaps an infinite number of points that can serve as $\pi(x)$. Continuing from the previous example, we can illustrate how this appears visually.

```
R> pi_x <- project_pi(P,x)
R> c <- 'blue'
R> cc <- 'red'
R> draw.tpolytope.3d(P,c,cc)
R> points(c(x[2],pi_x[2]),c(x[3],pi_x[3]),pch=19,col='green')
R> lines(c(x[2],pi_x[2]),c(x[3],pi_x[3]),lty='dashed')
```

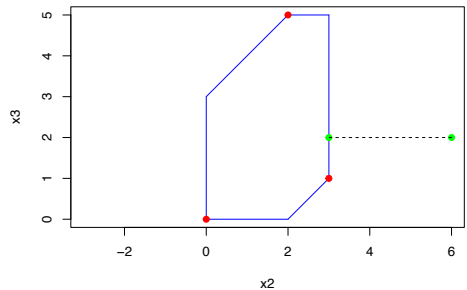

Figure 12: Projection of the point $x = (0, 6, 2)$ onto the tropical polytope $\mathcal{P}$, where $\mathcal{P} := \{(0, 0, 0), (0, 3, 1), (0, 2, 5)\}$. Here $\pi(x) = (0, 3, 2)$ but any point on the boundary of $\mathcal{P}$ from $(0, 3, 2)$ to $(0, 3, 1)$ could serve as $\pi(x)$.

The functions introduced in this section serve as standalone functions but they also provide the basis of other functions in the **TML** package. In the sections that follow, we introduce the statistical tools that leverage these basic functions.

### 2.3 Calculating a tropical Fermat-Weber point

In this section we introduce several functions that allow us to calculate tropical Fermat-Weber points. The output of these functions are incorporated in a variety of statistical methods included in the **TML** package, some of which are described below. Here we introduce two methods of finding the tropical FW point. The first is a method that uses linear programming while the second employs a fast gradient-based method.

The first method to find a tropical FW point for a given set of data is introduced in the `Trop_FW()` function. This function employs a linear programming approach to find a tropical FW point for a given set of data $V = \{v^1, \ldots, v^s\}$ by solving the constrained optimization problem

$$\min_{y} \quad \sum_{i=1}^{s} d_{tr}(v^i, y) \tag{12}$$

$$y_j - y_k - v_j^i + v_k^i \leq -d_{tr}(v^i, y) \; \forall \; i \in [s] \text{ and } 1 \leq j, k \leq e \tag{13}$$

$$y_j - y_k - v_j^i + v_k^i \geq d_{tr}(v^i, y) \; \forall \; i \in [s] \text{ and } 1 \leq j, k \leq e. \tag{14}$$

The example below shows how this is employed by using simulated data set consisting of 150 normalized points in the tropical projective torus which is found in the **TML** package as `Sim_points`. Note that `Trop_FW()` takes as input points which are normalized using the `normaliz.polytope()` function. The `Sim_points` is already normalized so normalization is not required. A plot of the output for the code below is shown in Figure 13.

```
R> set.seed(23)
R> V <- Sim_points
R> FW <- Trop_FW(V)
R> plot(V[,2],V[,3],pch=19,cex=.8,xlab="v2",ylab="v3",asp=1)
R> points(FW[2],FW[3],pch=19,col='red')
```

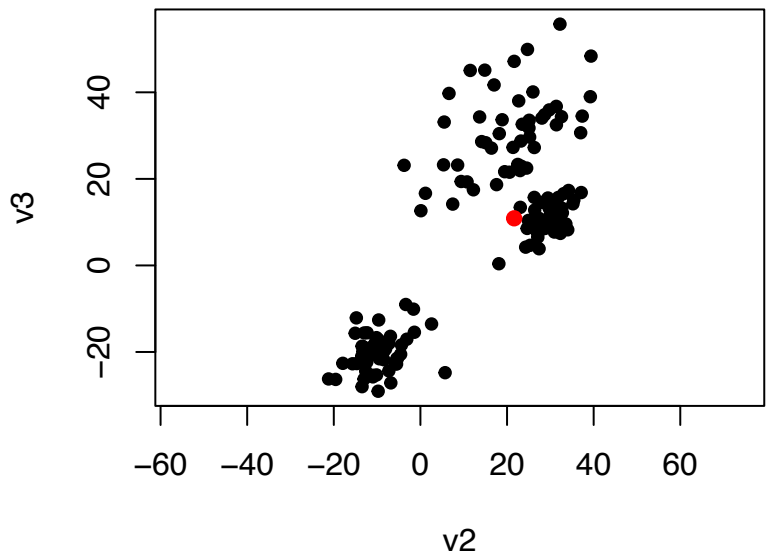

Figure 13: Tropical Fermat-Weber point calculated using the `Trop_FW()` function. Black points indicated the data. The red point is the tropical FW point.

While a FW point can be found directly using linear programming, gradient-based numerical methods like those developed and employed in [3] are much faster. Of particular interest in [3], is inferring the class of species tree associated with a set of gene tree. Because it is proven that the set of tropical FW points asymptotically converges to the maximum likelihood estimate (MLE) vectorised tree under this model this task can be reduced to finding a tropical FW point associated with the set of gene trees. [3] use FW points in lieu of MLE trees because the former is faster to compute numerically and it enjoys optimality sufficiency condition.

The following example illustrates how to compute a FW point using the gradient method of 1290 gene trees associated with a lungfish data set used in [18] by employing the function `FW_numerical()`. This data set is accessible in the **TML** package which includes a dissimilarity matrix when `lung_fish` is called and a vector of strings called `lf_labels` representing the species (or leaf) labels for the associated trees.

```
R> T <- lung_fish
R> labels <- lf_labels
```

```
R> omega <- FWpoint_numerical(T)
R> inferred_tree <- vector.to.equidistant.tree(omega)
R> inferred_tree$tip.label <- labels
R> plot(inferred_tree)
```

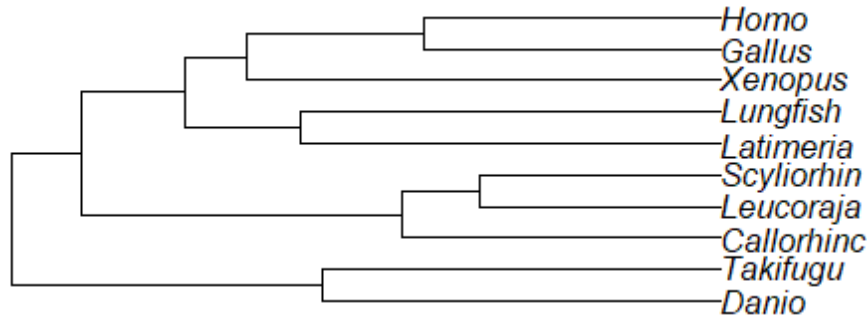

Figure 14: Equidistant tree representation of a Fermat-Weber point of 1290 gene trees from the lungfish data [18].

The functions introduced in this section serve as standalone functions but they also provide the basis of other functions in the **TML** package. In the sections that follow, we introduce the statistical tools that leverage these basic functions.

## 3  Tropical HAR as a main tool for inference

Markov chain Monte Carlo (MCMC) methods are an extremely important tool in statistical inference. Since their development in the first half of the twentieth century, MCMC methods have proven effective in a broad spectrum of scientific disciplines. Among the most flexible and easily constructed MCMC methods is the hit-and-run (HAR) sampler. Like all MCMC samplers, HAR samplers sample points according to a target distribution by moving from one point to another by defining a subset of a state space in terms of line segments. Both the current point and the possible next points fall on this line segment [30].

Up until recently, no MCMC samplers existed to sample points from a state space that could be defined as tropically convex. With the introduction of the sampler *HAR with extrapolation* as shown in [44], tropically convex sets can be sampled more effectively. As most methods in the **TML** rely on this novel HAR sampler, we devote this section to providing an in depth examination of its basic implementation as well as a number of variations.

## 3.1 HAR sampling from a tropical line segment

We begin by showing how we sample from a tropical line segment as the line segment is the basic geometric structure used in HAR sampling. There are two variations we illustrate here. The first shows how we sample points uniformly from a tropical line segment as shown in [44]. This method is implemented using the `HAR.TLineSeg()` function. For any two points in the tropical projective torus, we can define a tropical line segment and then sample from the line segment. The code that follows shows how to sample a single point from a tropical line segment.

```
R> set.seed(1)
R> u <- c(0,3,1)
R> v <- c(0,0,0)
R> BPs <- lapply(TLineSeg(u,v),function(x) normaliz.vector(x))
R> G_uv <- matrix(unlist(BPs), ncol = 3, byrow = TRUE)
R> draw.tpolytope.2d(G_uv,'red','blue')
R> pt<-normaliz.vector(HAR.TLineSeg(G_uv[1,],G_uv[nrow(G_uv),]))
R> points(pt[2],pt[3],pch=19,col='green')
```

To sample multiple points from a tropical line segment we can employ `HAR.TLineSeg()` in a `for()` loop in the following chunk. Figure 15 shows the building of the line segment, sampling a single point, and then sampling 200 points from the line segment.

```
R> u <- c(0,3,1)
R> v <- c(0,0,0)
R> poins <- matrix(0,nrow=200,3,TRUE)
R> for(i in 1:nrow(poins)){
    x <- HAR.TLineSeg(u,v)
    poins[i,] <- x
}
R> points(poins[,2],poins[,3],pch=19,cex=.3,col='green')
```

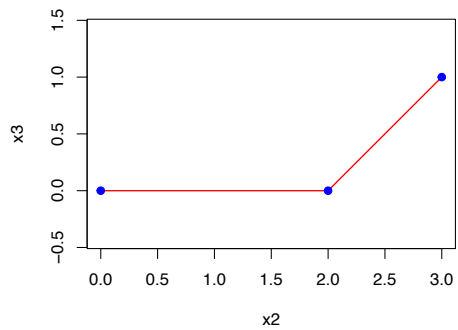

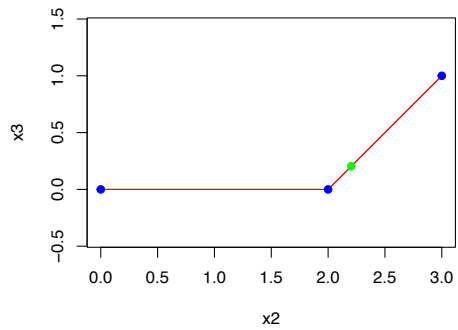

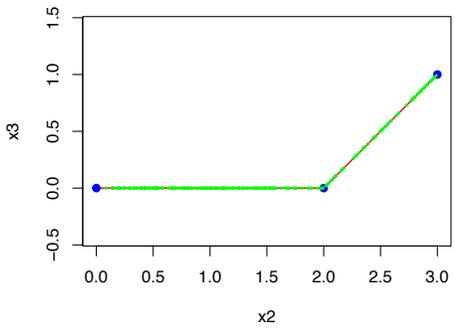

Figure 15: A tropical line segment (top) with the blue points indicating the break points and end points, a single point sampled from the tropical line segment in green (center), and 200 points sampled from the tropical line segment.

The **TML** package also gives the option to sample points from a tropical line segment about a point representing a centroid as shown in [4]. In practice, this is similar to Gaussian sampling about a point $\mu$ with a standard deviation $\sigma$ which controls dispersion. In the case of a tropical line segment, the scale parameter is based on tropical instead of Euclidean distance. The `HAR.TLineSeg.Norm()` function performs this calculation. Continuing from the previous example we show the results of sampling 200 points about a centroid represented by the point $\mu = (0, 2, 0)$ with a scale parameter $\sigma = 0.2$.

```
R> set.seed(1)
R> mu <- c(0,2,0)
R> sig <- .2
R> poins <- matrix(0,nrow=2000,3,TRUE)
R> for(i in 1:nrow(poins)){
    x <- HAR.TLineSeg.Norm(u,v,mu,sig)
    poins[i,]<-x
  }
R> u <- c(0,3,1)
R> v <- c(0,0,0)
R> BPs <- lapply(TLineSeg(u,v),function(x) normaliz.vector(x))
R> G_uv <- matrix(unlist(BPs), ncol = 3, byrow = TRUE)
R> draw.tpolytope.2d(G_uv,'red','blue')
R> points(poins[,2],poins[,3],pch=19,cex=.3,col='green')
```

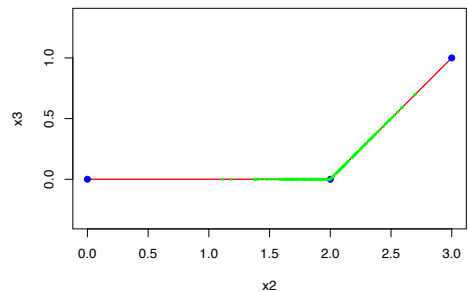

Figure 16: Sampling 200 points (green) about a centroid $\mu = (0, 2, 0)$ with scale parameter $\sigma = 0.2$.

Comparing the distance of the sampled points from the centroid $\mu$ we can determine the quantiles associated with the tropical distance.

```
R> dts <- apply(poins,1,function(x) trop.dist(mu,x))
R> quantile(dts)
        0%           25%           50%           75%          100%
0.0002291895 0.0636676116 0.1309480012 0.2323379661 0.7342599864
```

The quantiles associated with the tropical distance from the centroid according to a scale parameter $\sigma = 0.2$ is comparable to the Gaussian distribution in Euclidean space with a standard deviation equal to 0.2.

## 3.2 HAR sampling from a tropical polytope

Using the line sampling methods described above we can employ them to sample from tropical polytopes (see Definition 1.6). This is accomplished using the `TropicalPolytope.extrapolation.HAR()` function. This function allows the user to sample points uniforms from the $(e-1)-$trunk of a tropical simplex (Definition 1.20). Arguments for the `TropicalPolytope.extrapolation.HAR()` function include a matrix defined as the vertices of the tropical simplex, a initial point, and a scalar value indicating the number of intermediate points to sample between the initial state and the final state in the Markov chain. The state in the chain represents the sampled point. Figure 17 shows the results of sampling from a polytrope (top) and a generic tropical simplex (bottom).

```
R> set.seed(1)
R> P <- matrix(c(0,0,0,0,2,5,0,3,1),3,3,TRUE)
R> x0 <- c(0,2.5,3.2)
R> N <- 1000
R> poins <- matrix(0,nrow=N,ncol=ncol(P),TRUE)
R> for(i in 1:nrow(poins)){
    x <- TropicalPolytope.extrapolation.HAR(P,x0,I=50)
    x0 <- x
    poins[i,] <- x
   }
R> plot(poins[,2],poins[,3],xlab='x2',ylab='x3')
```

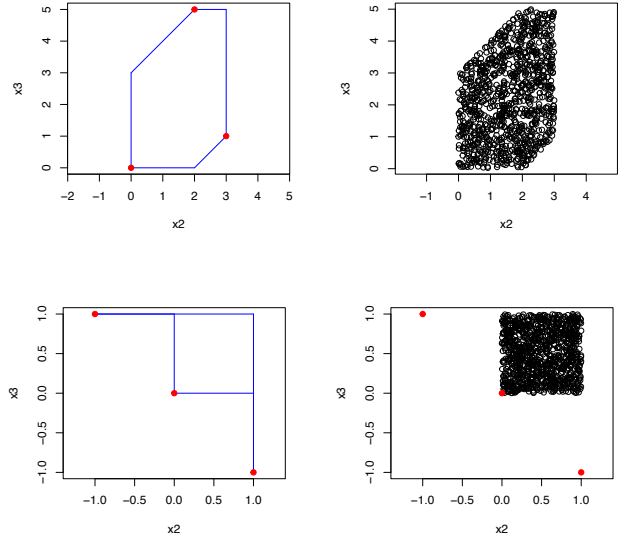

Figure 17: Two tropical polytroes, (left) with vertices indicated in red and results after using `TropicalPolytope.extrapolation.HAR()` to sample 1000 points from each polytope.

In addition, we can also sample points about a centroid (location parameter) with dispersion controlled by a scale parameter. This method is described in detail in [4]. In Euclidean space some HAR methods that sample points according to a Gaussian distribution do so by projecting the centroid onto the generated line and then sample about the projection according to a fixed standard deviation. In the tropical projective torus, the projection of a point onto a line segment is not usually unique. Therefore, sampling about a centroid in the tropical projective torus requires that we define the interval of possible projections of the centroid onto the line segment and sample uniformly from this interval. For detail on how this is accomplished see Chapter 2 in [4].

In the **TML** package, the `tropical.Gaussian()` function executes this method. Similarly, the `tropical.Gaussian()` function takes as inputs the vertices of a tropical simplex, a starting point, and a scalar value determining the length of the Markov chain. In addition, a point serving as a centroid representing the location parameter, and a scale parameter in the form of a scalar to control dispersion are also used. Figure 18 shows an output of sampling about a centroid.

```
R> set.seed(1)
R> P <- matrix(c(0,0,0,0,0,1000,0,1000,0),3,3,TRUE)
R> x0 <- c(0,2.5,3.2)
```

```
R> N <- 1000
R> poins <- matrix(0,nrow=N,ncol=ncol(P),TRUE)
R> mu <- c(0,500,500)
R> sd <- 4
R> for(i in 1:nrow(poins)){
    x <- tropical.gaussian(P,x0,I=50,mu,sd)
    x0 <- x
    poins[i,] <- x
   }
R> plot(poins[,2],poins[,3],xlab='x2',ylab='x3')
R> points(mu[2],mu[3],pch=19,col='red)
```

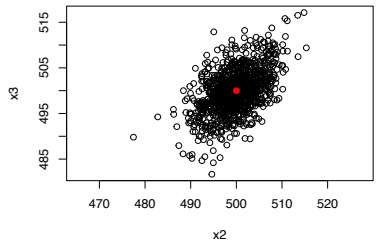

Figure 18: Sampling 1000 points about a centroid using the `tropical.gaussian()` function.

The functions in this section illustrate the basics of HAR sampling over the tropical projective torus. In the next section we provide specific data science applications using the methods in the **TML** package.

## 4 Machine learning applications

In this section we provide several applications of the methods available in the **TML** package. We begin with an application to show how to estimate the volume of a tropical polytope as described in [7], which is a NP-hard problem [12]. Next, we provide a supervised learning method involving tropical logistic regression applied to classifying phylogenetic trees as introduced in [2]. Then we show how to apply unsupervised learning in the form of tropical principal component analysis, again, applied to phylogenetic trees [25]. Finally, we show how to implement a non-parametric tropical kernel density estimation method as a way to identify outliers related to phylogenetic trees on $[m]$ leaves [43].

## 4.1 Application 1: Volume estimation of a tropical polytope

A challenging problem in polyhedral geometry is estimating the volume of polytopes and is no less challenging in the tropical setting. In this section, we follow the methods devised and illustrated in [7]. In general, for a given tropical polytope, $\mathcal{P}$, this involves finding a minimum enclosing tropical ball, denoted $B_r(\mathcal{P})$. By sampling from $B_r(\mathcal{P})$, which is of known volume, we can estimate the volume of $\mathcal{P}$ by multiplying the volume of $B_r(\mathcal{P})$ by the proportion of sampled points with membership in $\mathcal{P}$.

This application begins with computing $B_r(\mathcal{P})$ for a give $\mathcal{P}$ using the `min_enc_ball()` function. The output of the function is a two element list with the first element representing the center point of the ball and the second element representing the radius of the tropical ball in terms of tropical distance.

```
R> P <- matrix(c(0,0,0,0,3,1,0,2,5),3,3,TRUE)
R> B <- min_enc_ball(P)
R> B

[[1]]
[1] 0.0 2.0 2.5

[[2]]
[1] 2.5
```

Next we use the `trop_bal.vert()` to obtain the points in the minimum generating set of $B_r(\mathcal{P})$. The output is a matrix with with rows representing the points in the minimum generating set $V'$ of $B_r(\mathcal{P})$.

```
R> BR <- trop_bal.vert(B[[1]],B[[2]])

     [,1] [,2] [,3]
[1,]    0 -0.5  0.0
[2,]    0  4.5  2.5
[3,]    0  2.0  5.0
```

Using the output of points from the `trop_bal.vert()` function and the original tropical polytope, $\mathcal{P}$, we can estimate the volume of $\mathcal{P}$. This is accomplished through the use of the `Trop_Volume()` function. Inputs include an the matrix representing the tropical points defining the tropical ball, a matrix representing the points defining the original tropical polytope, an initial point, the number of points to sample, a scalar value representing the length of each Markov chain, and the radius, $r$, of $B_r(\mathcal{P})$.

```
R> set.seed(1)
R> x0 <- c(0,1.5,.4)
```

```
R> S <- 200
R> I <- 50
R> R <- B[[2]]
R> Trop_Volume(BR,P,x0,S,I,R)

    [[1]]
[1] 0.67

[[2]]
[1] 18.75

[[3]]
[1] 12.5625
```

The output of the `Trop_Volume()` function is a list containing three elements. The first element represents the proportion of points falling in the polytope of interest. The second element represents the volume of $B_r(\mathcal{P})$ and the third represents the volume estimate of the tropical polytope.

## 4.2  Application 2: Tropical logistic regression

We now introduce the supervised learning method of tropical logistic regression as applied to phylogenetics. In [3], tropical logistic regression is introduced and shown to outperform classical logistic regression when applied to phylogenetic trees. Specifically, tropical logistic regression is a binary classification method applied to a given set of phylogenetic trees. The classifications represent membership into one of two species trees.

Next, we turn to using tropical logistic regression to the problem of classifying gene trees according to the species tree that generated them. For this application, we use the two sets of 1000 phylogenetic trees, `Sim_Trees11` and `Sim_Trees21` (or simply $T1$ and $T2$ where the ratio $R = \frac{SD}{N_e} = 1$) where each phylogenetic tree has ten leaves and is represented as an ultrametric. Using $T1$ and $T2$, the task is now to classify unseen gene trees.

The tropical logistic regression model first infers the species tree that generated the corresponding trees of each class. Under the coalescent model which is explained in Section 1.4, the species trees and gene trees are equidistant and so the corresponding vectors are ultrametric. However, a Fermat-Weber point, which is used in lieu of the MLE tree, may not be an ultrametric. Ideally, we would like to have an additional constraint, requiring that the species tree be an ultrametric. By adding a regularization term that penalises deviations from the space of ultrametrics, we instead consider the modified Fermat-Weber point

$$\underset{\omega \in \mathbb{R}^e}{\arg\min} \left\{ \sum_{i=1}^{N} d_{\mathrm{tr}}(x_i, \omega) + \lambda \|\omega - \pi(\omega)\|^2 \right\},$$

where $x_i \in \mathbb{R}^e/\mathbb{R}\mathbf{1}$ is the $i$-th vectorized gene tree, $\pi$ is a projection onto the space of ultrametrics and the $\lambda$ is the regularization rate. This method is employed in the standalone function `FWpoint.num.w.reg()` which is incorporated in the tropical logistic regression method `trop.logistic.regression()`. The following example shows how tropical logistic regression can be employed.

```
R> library("ROCR")
R> D <- rbind(Sim_Trees15,Sim_Trees25)
R> Y <- c(rep(0,dim(Sim_Trees11)[1]), rep(1,dim(Sim_Trees21)[1]))
R> N <- length(Y)
R> set.seed(1)
R> train_set <- sample(N,floor(0.8 * N)) ## 80/20 train/test split
R> pars <- trop.logistic.regression(D[train_set,],
            Y[train_set], penalty=1e4)
R> test_set <- (1:N)[-train_set]
R> Y.hat <- rep(0, length(test_set))
R> for(i in 1:length(test_set))
        Y.hat[i] <- prob.class(pars, D[test_set[i],])
R> Logit.ROC <- performance(prediction(Y.hat, Y[test_set]),
                 measure="tpr", x.measure="fpr")
R> plot(Logit.ROC, lwd = 2,
        main = "ROC Curve for Logistic Regression Model")
R> AUC <- performance(prediction(Y.hat, Y[test_set]),
                        measure="auc")@y.values
R> AUC
[[1]]
[1] 0.970966
```

From the example above we see that the area under the curve associated with the receiver operator characteristic (ROC) curve is close to one. This indicates a near-perfect classification for the given example. Figure 21 provides a visual representation of the ROC curve associated with the example above.

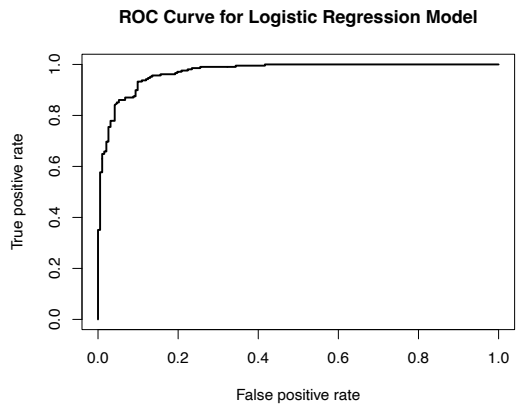

Figure 19: ROC curve produced by the code snippet above.

## 4.3 Application 3: Tropical PCA

Principal component analysis (PCA) is an unsupervised learning technique used for dimension reduction. Tropical PCA is no different but focuses on finding a best-fit tropical polytope for some data in the tropical projective torus. As in the previous section, we focus on the space of equidistant trees on $[m]$ leaves. As shown in the previous section, an equidistant tree can be defined as an ultrametric. In the **TML** package, tropical principal component analysis focuses on the tree space defined as the space of ultrametrics on $[m]$ leaves. This method was first introduced in [25] and extended in [41].

In the examples below we instead use simulated data where each point resides in the tropical projective torus. The best-fit polytope, specifically a tropical triangle, is calculated through the use of the `tropical.PCA.Polytope()` function. This function takes an iterative approach to finding the vertices of the best-fit tropical triangle by incorporating vertex HAR with extrapolation which was shown in Section 3.2. The primary purpose is to visualize the data along with the associated tropical triangle which is shown through the code that follows.

```
R> set.seed(1)
R> s <- 3 #number of vertices. Here it is a tropical triangle
R> d <- 3 ## dimension
R> N <- 100 ## sample size
R> V <- matrix(c(100, 0, 0, 0, 100, 0, 0, 0,
                100, -100, 0, 0, 0, -100,
                0, 0, 0, -100), 6, 3, TRUE)
R> D <- matrix(rep(0, N*d), N, d)
R> D[, 1] <- rnorm(N, mean = 5, sd = 5)
R> D[, 2] <- rnorm(N, mean = -5, sd = 5)
```

```
R> D[, 3] <- rnorm(N, mean = 0, sd = 5)
R> index <- sample(1:N, s)
R> S <- D[index,]
R> DD <- pre.pplot.pro(S, D)
R> for(i in 1:N)
R>   DD[i, ] <- normaliz.vector(DD[i, ])
R> res <- tropical.PCA.Polytope(S, D, V, I = 1000,50)
R> DD <- pre.pplot.pro(res[[2]], res[[3]])
R> trop.tri.plot.w.pts(normaliz.ultrametrics(res[[2]]), DD)
```

The output of this code provides the tropical triangle shown in Figure 20.

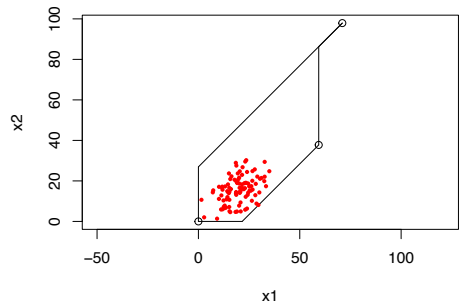

Figure 20: Best-fit tropical triangle found using `tropical.PCA.Polytope()`.

## 4.4 Application 4: Tropical kernel density estimation

A kernel density estimator (KDE) is a non-parametric density estimation method which uses kernel functions. This is a useful method in determining a number of data characteristics when the distribution of the data is unknown. This technique uses a kernel function, $\kappa(.)$ which is simply a non-negative, smooth function and conjunction with a bandwidth parameter [33, 37]. Like any density function however, there also must exist some normalizing constant, $C$ such that $\frac{1}{C}\kappa(.)$ integrates to one. A common kernel density estimator is the Gaussian kernel with mean zero and standard deviation equal to one

$$\frac{1}{\sqrt{2\pi}} \exp\left(-\frac{x^2}{2}\right). \qquad (15)$$

In 15, the normalizing constant is the fraction prior to the exponential function. The bandwidth, or dispersion, parameter is represented by the standard deviation in the case of the Gaussian kernel. Importantly, a reasonable density estimate requires an appropriate choice for bandwidth.

Kernel density estimation over the tropical projective torus has previously been investigated by Weyenberg et al. in [39], which specifically focused on kernel density estimation over the space of phylogenetic trees using what is called the BHV metric. One of the more challenging aspects of their method was that the location of the center of the kernel function causes the value of the normalizing constant to vary. This requires a recalculation of the normalizing constant for each data point.

As an alternative to this method In [43], Yoshida et al. introduced the notion of kernel density estimation over the treespace represented as the space of ultrametrics on $[m]$ leaves using the tropical metric in conjunction with a Laplacian kernel function. Conducted experiments suggested that the normalizing constant remains constant regardless of the center of the function with $m \geq 5$. Bandwidth, which is based on the tropical metric, is chosen using a "nearest neighbor" approach as in [38] meaning that the bandwidth parameter is equal to the tropical distance to the closest other data point.

The **TML** package provides tropical kernel density estimation in the form of the `tropical.KDE()` function. The method leverages two functions from the **KDETrees** package called `pw.trop.dist()` and `bw.nn()` to first calculate the pairwise tropical distance between each data point and then find the bandwidth parameter for each data point [38]. In [38], Weyenberg et al. show how, using their BHV metric-based approach, to identify outliers in a set of gene trees. In this case, an outlier tree is a tree that falls in the tail of the distribution of trees.

Yoshida et. al conducted a similar experiment using the method developed in [43] on a the space of ultrametrics on $m = 10$ leaves to identify such outliers with fixed effective populations but differing species depths (SD). The species depth indicates the number of epochs between the common ancestor of all species (root node) and present day (leaf nodes). For this experiment, we consider an effective population of $N_e = 100000$ and varying species depths such that we obtain a sequence of ratios, $R$, of SD to $N_e$ equal to 0.25, 0.5, 1, 2, 5, and 10.

The example below consists, again, of using the same two sets of 1000 simulated gene trees, $T1$ and $T2$, with ten leaves where $R = 5$ (represented as `Sim_Trees1` and `Sim_Trees2`). We then provide the the cumulative results for each $R$ in the plots of the receiver operator characteristic (ROC) curves for each $R$ that follow.

In order to determine how well we can identify ouliers using the `tropical.KDE()` function, we will examine each tree in $T2$ as it is appended to the set of trees in $T1$. Using the pairwise tropical distance function, `pw.trop.dist()`, and `bw.nn()`, we find the bandwidth value for each tree in the set. Then, we calculate the density value using `tropical.KDE()` function. We determine how well the method identifies outliers by examining the receiver operator characteristic (ROC) curve. The larger the value of the area under the ROC curve, the better the method is at identifying outliers. In the code chunk below, we assume the density estimate on the final trial for all trees with original membership in $T1$ is representative of density estimates from previous trials. Therefore, when calculating values for the ROC curve in the code chunk below, we only use those density estimates.

```
R> set.seed(1)
R> D1 <- Sim_Trees15
R> D2 <- Sim_Trees25
R> I <- 1000 ## The number of trials
R> Q5 <- rep(0, I)
R> N1<-nrow(D1)
R> for(i in 1:I){
        D <- rbind(D1, D2[i,])
        T <- dim(D)[1]
        P5 <- rep(0, T)
        X <- 1:T
        M <- pw.trop.dist(D, D)
        sigma <- bw.nn(M)
        P5 <- tropical.KDE(D, n, sigma, h = 2)
        Q5[i] <- P5[T]
        print(i)
    }
R> y <- c(rep(1, N1), rep(0, I))
R> predProbKDE5 <- c(P5[1:N1], Q5)
R> KDE5.ROC <- performance(prediction(predProbKDE5, y),
    measure="tpr", x.measure="fpr")
```

In general, we see that as $SD$, and therefore $R$, increases, so does the AUC value as shown below. When we reach the experiment representing $R = 10$, we see perfect classification, indicated by the AUC being equal to one. Figure 21 shows the associated ROC curves for each value of $R$. This provides us with a visual representation of the AUC values below.

```
R> KDE025.AUC <- performance(prediction(predProbKDE025, y),
                        measure=``auc'')@y.values
R> KDE025.AUC

[[1]]
[1] 0.563876

R> KDE05.AUC <- performance(prediction(predProbKDE05, y),
                        measure="auc")@y.values
R> KDE05.AUC

[[1]]
[1] 0.630703

R> KDE1.AUC <- performance(prediction(predProbKDE1, y),
                        measure="auc")@y.values
R> KDE1.AUC
```

```
[[1]]
[1] 0.697034

R> KDE2.AUC <- performance(prediction(predProbKDE2, y),
                            measure="auc")@y.values
R> KDE2.AUC

[[1]]
[1] 0.87902

R> KDE5.AUC <- performance(prediction(predProbKDE5, y),
                            measure="auc")@y.values
R> KDE5.AUC

[[1]]
[1] 0.998542

R> KDE10.AUC <- performance(prediction(predProbKDE10, y),
                             measure="auc")@y.values
R> KDE10.AUC

[[1]]
[1] 1
```

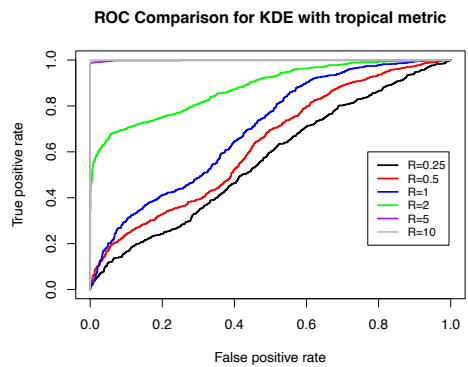

Figure 21: ROC curve for outlier detection when comparing two sets of gene trees with varying $SD$. Note that an increasing $R$ indicates an increasing $SD$.

# 5  Conclusion

This paper provides a basic description of the tropical machine learning methods and functionality of the **TML** package in R. While we provide a thorough descriptions of most available methods in the **TML** package we cannot cover everything. One important unsupervised method not covered are clustering methods over the tropical projective torus. We reccommend the read see [5] for a thorough treatment.

As shown in [24], all Euclidean statistical models can be described in terms of tropical algebra. With this in mind, we anticipate that `TML()` package will continue to be improved and expanded as new tropical data science methods are developed. We are already observing alternative methods of employing tropical support vector machines using HAR methods and neural networks in terms of tropical algebra. We encourage collaborators to provide input and recommendations via the **TML** GitHub page at https://github.com/barnhilldave/TML/issues.

# Acknowledgments

The authors would like to thank David Kahle for is input on the development of the **TML** package. RY and DB are partially supported from NSF DMS 1916037. GA is funded by EPSRC through the STOR-i Centre for Doctoral Training under grant EP/L015692/1. KM is partially supported by JSPS KAKENHI Grant Numbers JP22K19816, JP22H02364.